# Emergence of a phonological bias in ChatGPT

Juan Manuel Toro[1,2]
1. Institució Catalana de Recerca i Estudis Avançats (ICREA)
2. Universitat Pompeu Fabra

**This is a preliminary draft. For a peer-reviewed, updated manuscript reporting more recent data, please check**

Contact information:
Juan Manuel Toro
Universitat Pompeu Fabra
C. Ramon Trias Fargas, 25-27
CP 08005, Barcelona (Spain)
juanmanuel.toro@upf.edu

## Abstract

Current large language models, such as OpenAI's ChatGPT, have captured the public's attention because how remarkable they are in the use of language. Here, I demonstrate that ChatGPT displays phonological biases that are a hallmark of human language processing. More concretely, just like humans, ChatGPT has a consonant bias. That is, the chatbot has a tendency to use consonants over vowels to identify words. This is observed across languages that differ in their relative distribution of consonants and vowels such as English and Spanish. Despite the differences in how current artificial intelligence language models are trained to process linguistic stimuli and how human infants acquire language, such training seems to be enough for the emergence of a phonological bias in ChatGPT.

## Introduction

Recent advances in artificial intelligence have allowed the development of powerful large language models. OpenAI's ChatGPT is perhaps the one that has made a most notable impact since its presentation to the public. ChatGPT not only answers questions in a comprehensive manner, it can write college-level essays or short tv scripts. The elaborate responses produced by the chatbot have a striking similarity to how humans use language. Here, we explore if phonological biases that are widely observed in human language processing also emerge in current artificial intelligence language models.

In humans, acquiring a language leads to the emergence of processing biases. A well-known example is that of the consonant bias. This bias consists in a strong reliance on consonants over vowels for lexical access. The consonant bias has been extensively documented across different ages (from young infants to adults), modalities (oral, written), native languages (English, French, Spanish, Dutch) and tasks (word learning, word reconstruction, masked priming). From their second year of life, humans focus more heavily on the consonants than on the vowels to identify and learn new words (1) in a trend that is kept through adulthood. In a word reconstruction task, if adult listeners are acoustically presented with a non-sense word, such as “kebra”, and are asked to change only 1 phoneme to form a real word, they tend to keep intact the consonantal tier of the word and change a vowel (thus, forming the word “kobra”, instead of the word “zebra”; 2). Similarly, in a masked priming task, primes that keep the consonantal tier of the target word prime as much as primes identical of the target word. In contrast, primes that keep the vowel tier of the target word do not produce a reliable priming effect (3). Even more, statistical computations that are linked to the segmentation and learning of novel words are preferentially computed over consonants and not over vowels (4, 5). In humans, there is a thus a strong phonological bias that favors the use of consonants for the identification of words. Do current artificial large language models display a similar consonant bias?

## Results

We asked ChatGPT which of two non-words, one with a vowel change and one with a consonant change, was more similar to a target word (using the prompt "If you have to choose, which of these two non-words is more similar to “natural”, “nateral” or “nalural”?"; 6). Across 100 different words, the program exhibited a clear consonant bias, consistently choosing the non-word that left the consonantal tier of the target word intact over the non-word that preserved the vowel tier in up to 76% of the trials (C tier=76, V tier=24; Wilcoxon signed rank test $Z=1212$, $p<0.005$) independently of the length of the word, syllabic structure, position of the letter change or syntactic category of the word. In humans, the consonant bias is independent of native language (3). We tested ChatGPT in Spanish using the prompt “Si tienes que elegir, cual de estas dos no-palabras se parece más a [target word], [non-word 1] o [non-word 2]?”. We also observed a strong tendency to prefer non-words that kept the consonantal tier of the target word (C tier=74, V tier=26; $Z=1313$, $p<0.005$; see Figure 1). The observed evidence suggests that ChatGPT relies more heavily on consonants than on vowels to identify words.

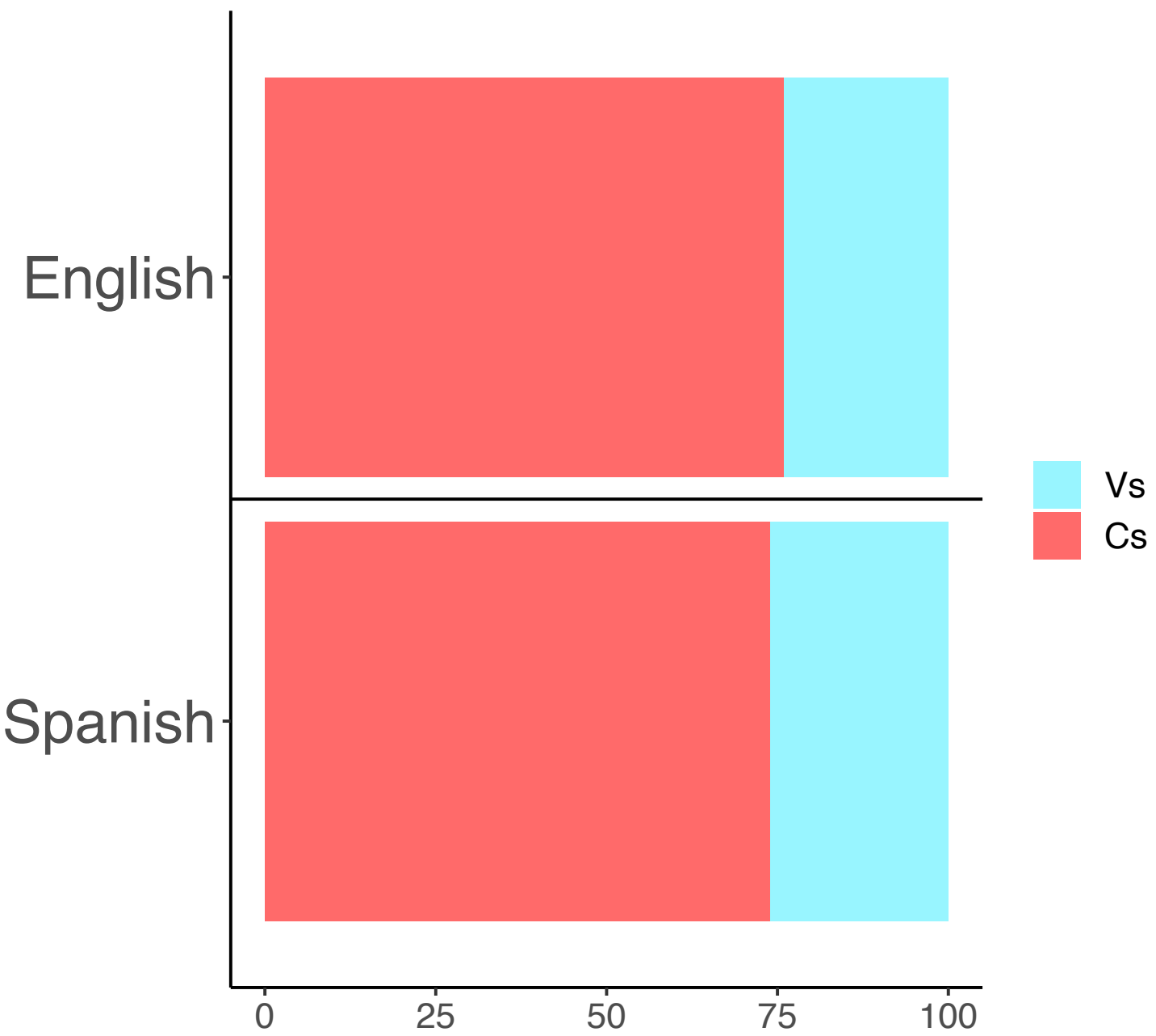

**Figure 1. A consonant bias in ChatGPT.** Number of trials in which ChatGPT chose non-words that kept the consonantal tier (Cs) over non-words that kept the vowel tier (Vs) as more similar to a target word.

**Discussion**

Natural languages tend to have more consonants than vowels. Out of 563 languages analyzed, not one has more vowels than consonants (7). English and Spanish, the two languages in which we tested the language model, differ in their relative distribution of consonants and vowels. While British English has 17 vowels and 24 consonants, Castilian Spanish has only 5 vowels and 20 consonants, with slight variations in both languages depending on dialect. Thus, listeners exposed to this asymmetry between consonants and vowels might learn to rely more on consonants to access the lexicon, as they are more informative to disambiguate between different words (8). However, this might not explain why natural languages tend to use more consonants than vowels in the first place (9). The presence of a consonant bias in ChatGPT across the two languages that we tested suggests that it arises independently of specific asymmetries between different phonological representations. It also provides a parallel with the observed consonant bias in native speakers of diverse natural languages such as English, Spanish, French or Danish. Importantly, this preference to use consonants over vowels for word identification is not built into current large language models, such as ChatGPT (10). It seems to be a property that emerges from the training that prepares these models to process language.

To produce comprehensive sentences, large language models are trained on massive amounts of data, so they detect statistical dependencies between words and learn to predict what word most likely follows another in a given context. It might thus be the case that, by tracking the relative differences in the distribution of consonants and vowels present in natural languages, large language models learn to focus on the consonants to disambiguate between possible lexical tokens. Several arguments have been made on how the training process in large language models contrasts with how infants acquire language (11,12). From a young age, infants generalize grammatical rules from relatively scarce experience and discover how to combine elements of language at different hierarchical levels (from the phonemes to the syllable and the word) to potentially create an infinite number of sentences. The original hypothesis about the consonant bias is that it facilitates language learning by establishing a "division of labor" between different phonological representations. While consonants mainly provide lexical information, vowels provide indexical information (13), and infants might benefit from this division of labor to facilitate lexical access. Interestingly however, just like humans, current large language models also seem to focus on consonants to identify words, drawing parallels between artificial and natural intelligences in their use of language (see also 14).

The evidence of a phonological bias in advanced chatbots is an example of emergent behavior in artificial intelligence. The fact that it parallels a phonological bias widely observed in human language processing, provides cues about the conditions under which some of the complexities of human language might arise across diverse cognitive systems.

## Methods

We asked ChatGPT which of two non-words was more similar to a target word. We used the prompt "If you have to choose, which of these two words is more similar to [target word], [non-word 1] or [non-word 2]?". As targets, we selected 100 words in English (6). Crucially, the words had different length (2, 3 and 4 syllables), syllabic structure (CV, CVC, CCV and VC) and belonged to different syntactic categories (nouns, verbs and adjectives). To create the non-words, we changed either 1 vowel or 1 consonant in each target word. The change could be placed in any syllable of the target word, with the exception of the first syllable. Across different words, the changes were implemented in the onset, nucleus or coda of syllables. Importantly, the 2 non-words presented in each trial had their vowel and consonant change implemented in the same syllable. That is, for a given target word, both non-words would have either the consonant or the vowel changed in the second syllable, third or fourth syllable. The order of presentation of the non-words within the prompt was balanced across trials. So, in half of the trials the non-word with a consonant change was presented first and in the other half of the trials the non-word with a vowel change was presented first. ChatGPT learns from previous interactions with the user, so no feedback was provided after its responses. Also, prompts were introduced one by one, so no automated script was used for their presentation.

We also queried ChatGPT in Spanish. The prompt was "Si tienes que elegir, cual de estas dos no-palabras se parece mas a [target word], [non-word 1] o [non-word 2]?". We selected as targets 100 words in Spanish (6). The creation and presentation of the non-words was the same as in English.

## Funding

This research was supported by grant PID2021-123973NB-I00 from the Ministerio de Ciencia e Innovación to JMT.

## Data availability

All data is available in the supplementary materials.

## Competing interests

The author declares no competing interests.

**Supplementary material**

**Emergence of a phonological bias in ChatGPT**

Juan Manuel Toro
ICREA – Universitat Pompeu Fabra
juanmanuel.toro@upf.edu

**Table 1.** List of 100 target words and non-words used for the English prompt "If you have to choose, which of these two non-words is more similar to [target word], [non-word 1] or [non-word 2]?". Non-words had either one consonant or one vowel change.

| Target | C change | V change |
|---|---|---|
| acid | acif | aced |
| adventure | adfenture | advonture |
| ambitious | ambilious | ambitaous |
| amicable | amifable | amicoble |
| argument | arsument | argiment |
| astronaut | asgronaut | astrinaut |
| attitude | attilude | attitade |
| avocado | avolado | avocudo |
| baby | maby | buby |
| barbecue | barnecue | barbocue |
| beautiful | beausiful | beautoful |
| bicycle | binycle | bicacle |
| brilliant | brilliart | brilliunt |
| calculate | calgulate | calcelate |
| calendar | camendar | calondar |
| camera | calera | camura |
| category | catelory | categary |
| chocolate | chocorate | chocolite |
| clever | clefer | clevor |
| compassionate | comnassionate | compossionate |
| computer | comluter | compater |
| confident | conrident | confodent |
| construction | conspruction | constrection |
| conversation | condersation | convirsation |
| design | delign | desegn |
| dinosaur | dilosaur | dinesaur |
| disparity | disnarity | dispurity |
| diva | dina | divu |
| document | docufent | docament |
| economy | ecoromy | econimy |
| elegant | elegart | elegunt |
| elephant | emephant | elophant |
| energetic | enedgetic | enurgetic |
| even | evef | evun |
| exercise | exencise | exurcise |
| explore | explome | explora |
| father | fatheg | fathor |
| flower | flomer | flowir |
| formidable | formisable | formidoble |
| generate | genesate | generote |
| generous | gederous | genarous |
| honest | honent | honust |
| hospital | hosmital | hosputal |
| imaginative | imapinative | imagunative |
| improve | imfrove | imprave |
| inevitable | inegitable | inevutable |
| infection | infeltion | infuction |
| insert | insent | insart |
| intelligent | inmelligent | intolligent |
| investigate | invesligate | investugate |
| island | islard | islond |
| keyboard | keymoard | keybeard |
| lemon | leton | lemin |
| library | ligrary | librory |
| listen | lisfen | listun |
| logical | lopical | logacal |
| magic | magil | magec |
| manage | mapage | manige |
| mansion | mansiod | mansian |
| metaphor | meraphor | metiphor |
| microphone | microsone | microphune |
| modest | modert | modost |
| mysterious | myslerious | mystorious |
| natural | nalural | nateral |
| notebook | nosebook | notubook |
| obsolete | obdolete | obsilete |
| open | opes | opun |
| optimistic | optimiltic | optimostic |
| organize | ordanize | orgunize |
| paper | pamer | papor |
| parent | parelt | parunt |
| passionate | passiodate | passionite |
| powerful | powenful | poworful |
| relax | relan | relix |
| river | rives | rivur |
| romantic | romaltic | romintic |
| satisfaction | satislaction | satisfoction |
| sensitive | sensipive | sensitove |
| sincere | sintere | sincure |
| social | sociam | sociul |
| study | sfudy | stidy |
| table | tabre | tabli |
| telephone | teledone | telephine |
| tomato | topato | tomuto |
| travel | traved | travul |
| triangle | triasgle | triengle |
| trustworthy | trustwopthy | trustwerthy |
| ultimate | ultisate | ultimote |
| umbrella | umfrella | umbrilla |
| understand | undepstand | undorstand |
| universal | unitersal | univorsal |
| universe | upiverse | unaverse |
| vanilla | vasilla | vanulla |
| vegetable | vegemable | vegetible |
| victory | vicpory | victary |
| vivacious | vipacious | vivecious |
| volunteer | volulteer | volanteer |
| write | wrine | writu |
| yesterday | yestenday | yesturday |
| zebra | zetra | zebru |

**Table 2.** List of 100 target words and non-words used for the Spanish prompt “Si tienes que elegir, cual de estas dos no-palabras se parece más a [target word], [non-word 1] o [non-word 2]?”. Non-words had either one consonant or one vowel change.

| Target | C change | V change |
|---|---|---|
| abierto | abiento | abiorto |
| abrazo | acrazo | abrizo |
| aburrido | amurrido | aberrido |
| agua | alua | agia |
| alegre | adegre | aligre |
| alegre | amegre | alugre |
| alto | almo | alte |
| alud | alun | aled |
| animal | animad | animol |
| aprender | apresder | aprunder |
| árbol | árbon | árbil |
| arco | arfo | arcu |
| armario | arbario | armurio |
| balón | balós | balín |
| bocadillo | bonadillo | bocudillo |
| bolígrafo | bosígrafo | bolágrafo |
| bonito | bosito | bonuto |
| brillante | brillacte | brillente |
| caballo | calallo | cabillo |
| calabaza | calaraza | calabiza |
| calcetín | caldetín | calcitín |
| caldero | calmero | caldiro |
| calendario | calesdario | calondario |
| cálido | cáfido | cáludo |
| caliente | calieste | calionte |
| caminar | cabinar | camunar |
| carrusel | carrunel | carrusul |
| casa | cafa | casu |
| celebrar | cepebrar | celobrar |
| cerrado | cerrafo | cerradi |
| césped | césred | céspud |
| cocinar | cocisar | cocinur |
| comer | coter | comor |
| comercio | comencio | comurcio |
| compartir | compastir | compurtir |
| complicado | complifado | complicudo |
| concentración | concenpración | concentrución |
| concierto | conciepto | conciurto |
| conectar | conentar | conictar |
| conservar | consenvar | consurvar |
| construir | constluir | constreir |
| decidir | detidir | decedir |
| delgado | delmado | delgodo |
| descubrir | desdubrir | descabrir |
| despertar | despentar | despurtar |
| difícil | dibícil | diféciI |
| dirigir | diribir | dirigar |
| divertido | divestido | divurtido |
| dormida | dorpida | dormoda |
| enseñar | enkeñar | ensiñar |
| épico | énico | épuco |
| escribir | esclibir | escrobir |
| escritorio | escrilorio | escritario |
| escuchar | esnuchar | escochar |
| estudiar | estumiar | estuduar |
| filtro | filpro | filtri |
| flor | flon | flir |
| fuerte | fuerde | fuerti |
| grande | granfe | grandu |
| hablar | hablaf | hablir |
| hermoso | herdoso | hermeso |
| horrible | horribre | horriblu |
| hospital | hosfital | hospatal |
| insecto | inselto | insocto |
| instante | insfante | instunte |
| interesante | intenesante | interosante |
| investigar | invesfigar | investugar |
| invierno | inviesno | inviurno |
| iris | irin | ires |
| jirafa | jinafa | jirofa |
| joven | jovel | jovun |
| manzana | mandana | manzina |
| marmota | marlota | marmeta |
| mejora | mebora | mejera |
| nadar | nadap | nadur |
| nariz | narip | naruz |
| niña | nifa | niñu |
| observar | obselvar | obsirvar |
| oler | ober | olir |
| opaco | oñaco | opico |
| ordenador | orfenador | ordunador |
| pantalón | pandalón | pantilón |
| pequeño | pequedo | pequeñe |
| primero | prifero | primuro |
| rápido | rálido | rápudo |
| reparar | relarar | reporar |
| satélite | samélite | satúlite |
| segundo | segurdo | segendo |
| sencillo | sentillo | sencello |
| sílaba | sígaba | síloba |
| sombrilla | somgrilla | sombrolla |
| soñar | sodar | soñir |
| tambor | tambos | tambar |
| trabajar | tradajar | trabojar |
| tranquilo | tranquipo | tranquile |
| triángulo | triásgulo | trióngulo |
| universo | univetso | univurso |
| valiente | valierte | valionte |
| visitar | vinitar | visatar |
| zapato | zadato | zapeto |